\documentclass[10pt,twocolumn,letterpaper]{article}

\usepackage[pagenumbers]{cvpr} 

\usepackage{xspace}
\usepackage{subcaption}
\usepackage{graphicx}
\usepackage[autolanguage]{numprint}
\usepackage[utf8]{inputenc}
\usepackage{microtype}
\usepackage{soul}
\usepackage{float}
\usepackage{textcomp, gensymb}
\usepackage{amssymb}
\usepackage{booktabs}
\usepackage{array}
\usepackage{relsize}
\usepackage[export]{adjustbox}
\usepackage[accsupp]{axessibility}

\definecolor{cvprblue}{rgb}{0.21,0.49,0.74}

\usepackage{caption}
\usepackage[linesnumbered,ruled,vlined]{algorithm2e}

\usepackage{footmisc}
\usepackage[dvipsnames]{xcolor}

\newcommand\blfootnote[1]{%
\begingroup 
\renewcommand\thefootnote{}\footnote{#1}%
\addtocounter{footnote}{-1}%
\endgroup 
}
\usepackage[pagebackref,breaklinks,colorlinks,citecolor=cvprblue]{hyperref}

\def\paperID{8364} 
\def\confName{CVPR}
\def\confYear{2025}

\title{MeshAnything V2: Artist-Created Mesh Generation\\ with Adjacent Mesh Tokenization}

\author{
\textbf{Yiwen Chen$^{1, 2}$} \quad 
\textbf{Yikai Wang$^{3\text{\textbf{\textcolor{red}{*}}}}$} \quad 
\textbf{Yihao Luo$^4$} \quad
\textbf{Zhengyi Wang$^{2, 3}$} \\  \quad \textbf{Zilong Chen$^{2, 3}$} \quad
\textbf{Jun Zhu$^{2, 3}$}  \quad  \quad 
\textbf{Chi Zhang$^{5\text{\textbf{\textcolor{red}{*}}}}$} \quad 
\textbf{Guosheng Lin$^{1\text{\textbf{\textcolor{red}{*}}}}$}
\vspace{0.2cm} \\
$ ^1$Nanyang Technological University \quad
$ ^2$Shengshu \\
$ ^3$Tsinghua University \quad
$ ^4$Imperial College London \quad $ ^5$Westlake University \\ 
{\fontsize{10}{10}\selectfont \url{https://buaacyw.github.io/meshanything-v2/}}
}

\begin{document}
\maketitle
\begin{figure*}[t!]
  \centering
   \includegraphics[width=\textwidth]{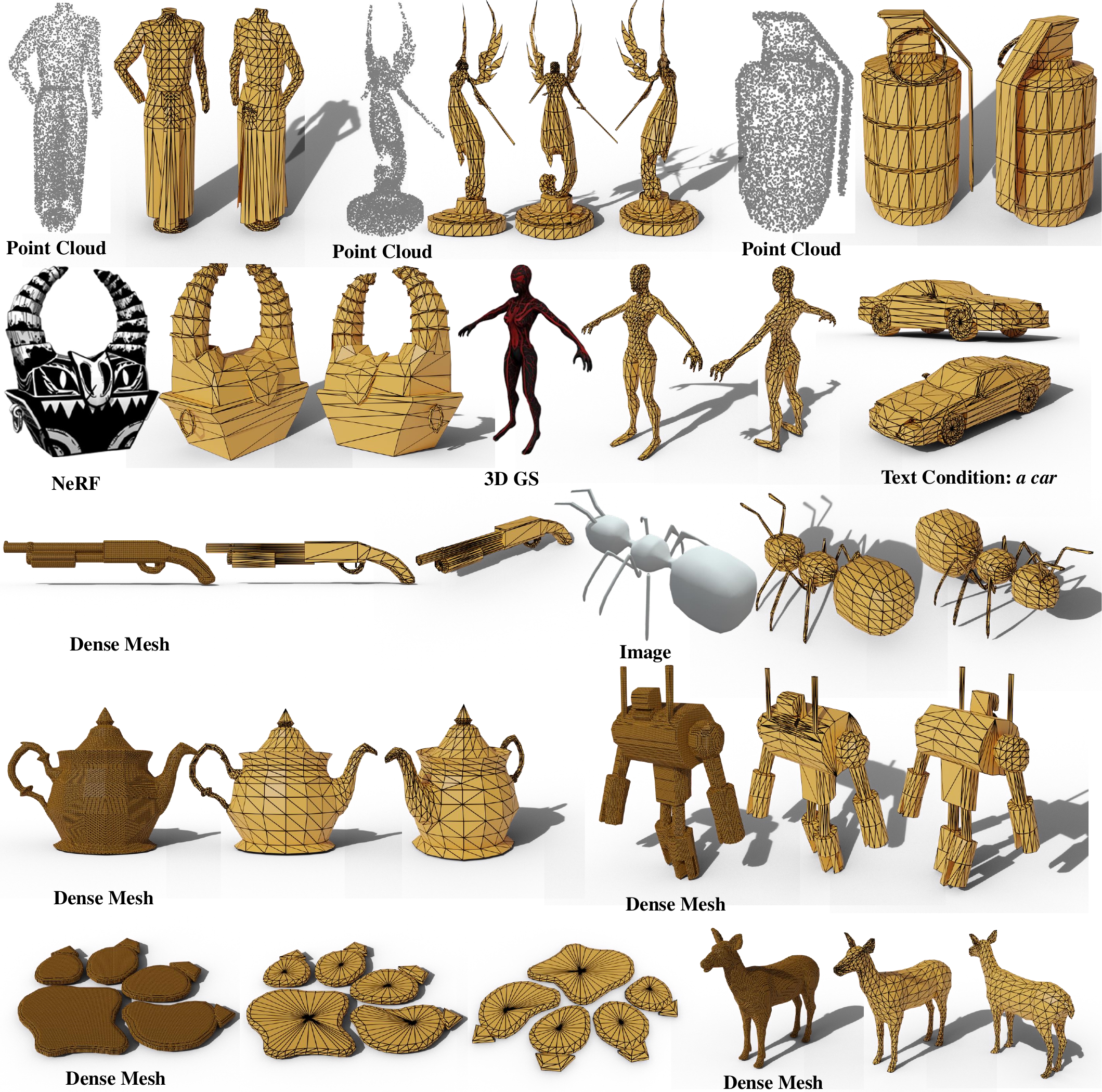}
   \caption{
\textbf{Equipped with the newly proposed Adjacent Mesh Tokenization (AMT), MeshAnything V2 significantly surpasses MeshAnything~\cite{chen2024meshanything} in both performance and efficiency.} MeshAnything V2 generates Artist-Created Meshes (AM) up to $1600$ faces aligned with given shapes. Combined with various 3D asset production pipelines, it efficiently achieves high-quality, highly controllable AM generation.}

\vspace{-3mm}

   \label{fig:teaser}

\end{figure*}

\begin{abstract}

Meshes are the de facto 3D representation in the industry but are labor-intensive to produce. Recently, a line of research has focused on autoregressively generating meshes. This approach processes meshes into a sequence composed of vertices and then generates them vertex by vertex, similar to how a language model generates text. These methods have achieved some success but still struggle to generate complex meshes. One primary reason for this limitation is their inefficient tokenization methods.  To address this issue, we introduce MeshAnything V2, an advanced mesh generation model designed to create Artist-Created Meshes that align precisely with specified shapes. A key innovation behind MeshAnything V2 is our novel Adjacent Mesh Tokenization (AMT) method. Unlike traditional approaches that represent each face using three vertices, AMT optimizes this by employing a single vertex wherever feasible, effectively reducing the token sequence length by about half on average. This not only streamlines the tokenization process but also results in more compact and well-structured sequences, enhancing the efficiency of mesh generation. With these improvements, MeshAnything V2 effectively doubles the face limit compared to previous models, delivering superior performance without increasing computational costs. We will make our code and models publicly available.

\end{abstract}    
\blfootnote{\textcolor{red}{*} Corresponding authors.}
\section{Introduction}

Due to the controllable and compact advantages of meshes, they serve as the predominant 3D representation in various industries, including games, movies, and virtual reality. For decades, the 3D industry has relied on human artists to manually create meshes, a process that is both time-consuming and labor-intensive. 

To address this issue, very recently, a line of work~\cite{nash2020polygen,alliegro2023polydiff,siddiqui2023meshgpt,weng2024pivotmesh,chen2024meshxl,chen2024meshanything} has focused on automatically generating Artist-Created Meshes (AMs) to replace manual labor. Inspired by the success of large language models (LLMs), these approaches treat AMs as sequences of faces and learn to generate them with autoregressive transformers~\cite{vaswani2017attention} in a manner similar to LLMs. Unlike methods that produce dense meshes in a reconstruction manner, these methods learn from the distribution of meshes created by human artists, thereby generating AMs that are efficient, beautiful, and can seamlessly replace manually created meshes.

Although these methods have achieved some success, they still face significant challenges. One major limitation is that current methods~\cite{nash2020polygen,alliegro2023polydiff,siddiqui2023meshgpt,weng2024pivotmesh,chen2024meshxl,chen2024meshanything} cannot generate meshes with a large number of faces. A primary reason for this constraint is the inadequacy of existing mesh tokenization methods. These methods treat a mesh as a sequence of faces, where each face consists of three vertices, and each vertex typically requires three tokens to represent. Consequently, each mesh is tokenized into a sequence nine times the number of its faces, resulting in substantial computational and memory demands. Besides, the resulting token sequence is highly redundant, which harms sequence learning and reduces performance.

Observing the above issues, we focus on the tokenization method of mesh generation in this work, aiming to improve the efficiency and quality of mesh generation. Extensive research in language models has demonstrated the importance of tokenization in sequence learning~\cite{wu2016google, sennrich2015neural, kudo2018sentencepiece}. Moreover, unlike text, which inherently has a sequential structure, meshes are graph-based structures with 3D characteristics. For any given mesh, there are countless ways to represent it as a 1D token sequence, making the influence of tokenization methods even more pronounced for meshes. Therefore, we emphasize that research on mesh tokenization methods is of great importance.
 
The impact of mesh tokenization on autoregressive mesh generation can be considered in two main aspects. The first is efficiency: representing a mesh with shorter, more compact token sequences leads to reduced context length, thereby decreasing memory and computational complexity. The second aspect is the regularity of the token sequence. A shorter token sequence is not always better for mesh generation; the regularity and pattern consistency of the sequence are crucial for effective sequence learning. Effective mesh tokenization must balance both efficiency and regularity to achieve high-quality, efficient mesh generation.

Considering the above, we introduce MeshAnything V2, an advanced model for generating Artist-Created Meshes. This model incorporate several key improvements to boost both performance and efficiency.
At the core of MeshAnything V2 is our novel Adjacent Mesh Tokenization (AMT) method. AMT optimizes the tokenization process by representing each face with a single vertex rather than the traditional three. As illustrated in Fig.~\ref{alg:AMT} and Algo.~\ref{alg:AMT}, AMT encodes adjacent faces using just one additional vertex, largely reducing the sequence length. When an adjacent face cannot be identified, a special token '\&' is introduced to indicate this interruption, allowing the model to resume from an unencoded face.

In previous mesh generation works, users were unable to control the number of faces generated by the model, often resulting in meshes that did not meet application requirements. To address this, we introduce a face count condition, allowing users to specify an approximate number of faces, ensuring that generated meshes align with desired specifications. Additionally, to enhance the robustness of AMT during inference, we incorporate Masking Invalid Predictions~\cite{nash2020polygen} to prevent the model from producing invalid tokens, such as generating another '\&' token immediately after an '\&' token.

Extensive experiments on the Objaverse~\cite{deitke2023objaverse} dataset demonstrate that AMT can halve the sequence length on average. This reduction translates to nearly a fourfold decrease in computational load and memory usage within the attention block. Furthermore, AMT remains effective across various mesh generation settings, including unconditional and conditional approaches~\cite{siddiqui2023meshgpt}, even when using VQ-VAE~\cite{van2017neural}. Finally, with the integration of AMT, MeshAnything V2 doubles the maximum number of faces generated compared to previous models, significantly boosting both performance and efficiency.

In summary, our contributions are:
\begin{enumerate}
    \item We introduce MeshAnything V2, a novel Artist-Created Mesh Generation model. V2 doubles the maximum number of faces that can be generated while achieving significantly better accuracy and efficiency.
    \item The core of V2 is our newly proposed mesh tokenization method, Adjacent Mesh Tokenization (AMT). Compared to previous tokenization methods, AMT requires approximately half the token sequence length to represent the same mesh, thereby fundamentally reducing the computational burden of Artist-Created Mesh generation.
    \item We conduct extensive experiments on various mesh tokenization methods to explore the essential properties required for effective mesh tokenization. Our experiments demonstrate that AMT significantly improves the efficiency and performance of mesh generation, and that the token sequence produced by mesh tokenization methods must balance compactness and regularity.

\end{enumerate}

\begin{algorithm}
\caption{Adjacent Mesh Tokenization (AMT)}
\label{alg:AMT}
\KwIn{$\mathcal{M}$: a triangle mesh}
\KwOut{A token sequence $Seq$ that represents $\mathcal{M}$}
Sort the vertices of $\mathcal{M}$ in ascending order by their coordinates\;
Sort the faces of $\mathcal{M}$ in ascending order by their vertex indices\;
Initialize an empty token sequence $Seq$\;
Initialize a face list $UnvisitedFaces$ as the faces of $\mathcal{M}$\;
Remove the first face from $UnvisitedFaces$\;
Append the three vertices of the removed face to $Seq$\;
Define a SpecialToken \& to indicate break signal. 

\While{$UnvisitedFaces$ is not empty}{
    \eIf{the last token in $Seq$ is \&}{
        Remove the first face from $UnvisitedFaces$\;
        Append the three vertices of the removed face to $Seq$\;
    }{
        $AdjacentVertices \gets$ vertices adjacent to both the last two vertices in $Seq$\;
         Filter out vertices from $AdjacentVertices$ that form faces with the last two vertices in $Seq$ that are not in $UnvisitedFaces$\;
        \If{$AdjacentVertices$ is not empty}{
           Sort $AdjacentVertices$ in ascending order by their coordinates\;
           Append the first vertex in $AdjacentVertices$ to $Seq$\;
        }\Else{
            Append \& to $Seq$\;
        }
    }
}

\Return $Seq$\;
\end{algorithm}

\section{Related Works}
\label{sec:related}

\begin{figure*}[t!]
    \centering
    \includegraphics[width=\textwidth]{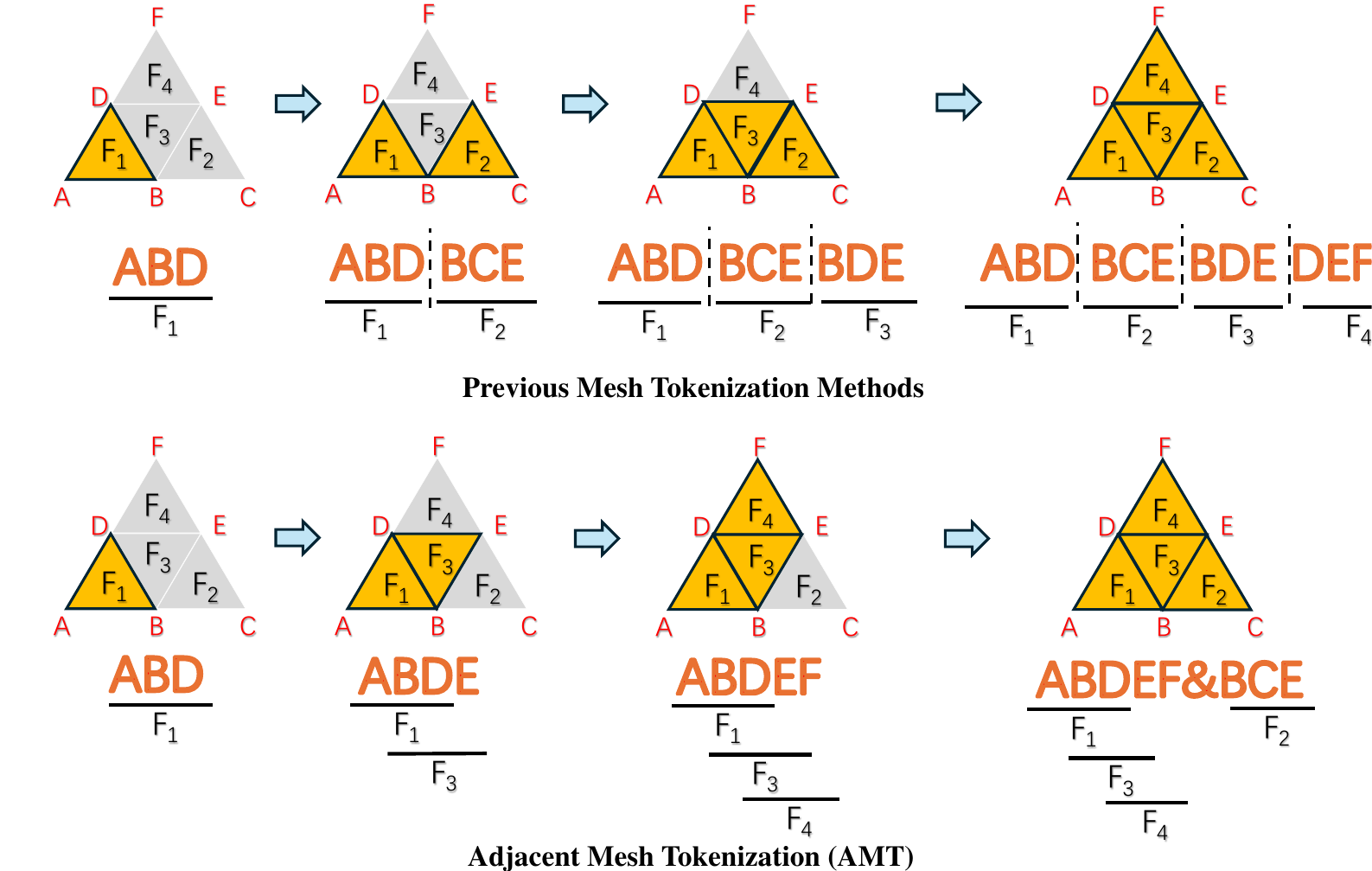}
\caption{
\textbf{Illustration of Adjacent Mesh Tokenization (AMT).} Unlike previous methods that use three vertices to represent a face, AMT uses a single vertex whenever possible. When this is impossible, AMT adds a special token \& and restarts. Our experiments demonstrate that AMT reduces the token sequence length by half on average. Its compact, and well-structured sequence representation enhances sequence learning, thereby significantly improving both the efficiency and performance of mesh generation.
}
    \label{fig:amt}
\end{figure*}

\subsection{Artist-Created Mesh Generation}
Diverging from previous works that produce dense meshes, recent works have focused on generating meshes created by human artists, i.e., Artist-Created Meshes (AMs)~\cite{nash2020polygen,alliegro2023polydiff,siddiqui2023meshgpt,chen2024meshxl,weng2024pivotmesh,chen2024meshanything}. These methods process meshes into ordered face sequences and learn to generate this sequence. \cite{nash2020polygen} first proposed using autoregressive transformers to sequentially generate vertices and faces. \cite{siddiqui2023meshgpt} use VQ-VAE to learn a mesh vocabulary and then learn this vocabulary with a decoder-only transformer. \cite{alliegro2023polydiff} differ from other methods by using a discrete diffusion model to generate AMs instead of an autoregressive transformer. \cite{chen2024meshxl} propose directly using the discretized coordinates of the vertex as the token index, bypassing the need for VQ-VAE as in~\cite{siddiqui2023meshgpt}. \cite{weng2024pivotmesh} use pivot vertices as a coarse mesh representation and then generate the complete mesh tokens. \cite{chen2024meshanything} generate AMs aligned with given shapes, which can be integrated with various 3D asset production methods to convert their results into AMs.

As shown in Fig.~\ref{fig:amt}, all of these methods process meshes into face sequences and use three vertices to represent a single face, resulting in highly redundant representations. Different from these methods, our newly proposed Adjacent Mesh Tokenization (AMT) uses a single vertex to represent a single face, providing a more compact and well-structured mesh representation, thereby significantly improving the efficiency and performance of mesh generation.

\subsection{3D Generation}
In recent years, 3D generation has gradually become one of the mainstream research directions in the field of 3D research. This area focuses on generating diverse, high-quality 3D assets for the 3D industry. Generative Adversarial Networks (GANs)~\cite{wu2016-3dgan,achlioptas2018learning,goodfellow2020generative} produce synthetic 3D data by training a generator and a discriminator network to distinguish between generated and real data. Very recently, a new line of works~\cite{hong2023lrm,liu2024one,shi2023mvdream,li2023instant3d,tang2024lgm,wang2024crm,tochilkin2024triposr,wei2024meshlrm,xu2024instantmesh} directly generate 3D assets in a feed-forward manner. \cite{hong2023lrm} pioneer these methods and use a transformer to directly regress the parameters of 3D models given conditions. Besides, applying diffusion models~\cite{ho2020ddpm} to directly generate 3D assets has also been widely researched~\cite{zhou2021pvd,nichol2022point-e,alliegro2023polydiff,lyu2023controllable,liu2023meshdiffusion,zhang2024clay}. \cite{zhang2024clay} lead the SOTA of current 3D generation methods by first generating high-quality 3D shapes with DiT and then producing detailed textures with material diffusion models.

As mesh is a crucial component in 3D generation, Artist-Created Mesh Generation~\cite{nash2020polygen,alliegro2023polydiff,siddiqui2023meshgpt,chen2024meshxl,weng2024pivotmesh,chen2024meshanything} is closely related to 3D generation. However, it differs significantly from previous 3D generation methods as it mainly focuses on sequence learning, which is rarely seen in other 3D generation methods, to produce high-quality mesh topology.

\section{Method}
\label{sec:method}

In this work, we introduce a novel mesh tokenization method, named Adjacent Mesh Tokenization (AMT). Compared to previous tokenization methods, AMT reduces the token sequence length by approximately half to represent the same mesh, thereby significantly lowering the computational overhead of mesh generation. AMT is detailed in Section~\ref{sec:AMT}. 

Next, in Section~\ref{sec:V2}, we integrate AMT into MeshAnything~\cite{chen2024meshanything} and introduce MeshAnything V2. 

\subsection{Adjacent Mesh Tokenization}
\label{sec:AMT}

In this section, we detail Adjacent Mesh Tokenization (AMT), a novel tokenization method for Artist-Created Mesh (AM) generation. Compared to previous methods, AMT processes the mesh into a more compact and well-structured token sequence by representing each face with a single vertex whenever possible. For simplicity, we describe AMT on triangle mesh. But it is worth noting that AMT can be easily generalized to the generation of meshes with variable polygons.

Tokenization is a crucial part of sequence learning, as it processes various data formats, such as text, images, and audio, into token sequences. The processed tokens are then used as ground truth inputs for training the sequence model. During inference, the sequence model generates a token sequence that is subsequently detokenized into the target data format. Therefore, tokenization plays a vital role in sequence learning, determining the quality of the data sequence that the sequence model learns from.

We first illustrate the tokenization methods used in previous methods~\cite{nash2020polygen,alliegro2023polydiff,siddiqui2023meshgpt,chen2024meshxl,weng2024pivotmesh,chen2024meshanything}. Although there are slight differences in detail, the previous tokenization methods can be unified as follows: Given a mesh $\mathcal{M}$, vertices are first sorted in ascending order based on their z-y-x coordinates, where z represents the vertical axis. Next, faces are ordered by their lowest vertex index, then by the next lowest, and so on. The mesh is then viewed as an \textbf{ordered} sequence of faces:

\begin{equation}
\label{Mesh2Face}
    \mathcal{M} := (f_1, f_2, f_3, \ldots, f_N),
\end{equation}

where $f_i$ represents the $i$-th face in the mesh, and $N$ is the number of faces in $\mathcal{M}$.

Then, each $f_i$ is represented as an ordered sequence of three vertices $v$:
\begin{equation}
\label{Face2V}
    f_i := (v_{i1}, v_{i2}, v_{i3}),
\end{equation}
where $v_{i1}, v_{i2},$ and $v_{i3}$ are the vertices that form the $i$-th face $f_i$ in the mesh. It is worth noting that $v_{i1}, v_{i2},$ and $v_{i3}$ have already been sorted and have a fixed order.

Substituting Equation~\eqref{Face2V} into Equation~\eqref{Mesh2Face} gives:
\begin{equation}
\label{Mesh2V}
    \mathcal{M} := ((v_{11}, v_{12}, v_{13}), \ldots, (v_{N1}, v_{N2}, v_{N3})) = \text{Seq}_V
\end{equation}
Due to the sorting, the resulting $\text{Seq}_V$ is unique and its length is three times the number of faces in the mesh. It is evident that $\text{Seq}_V$ contains a significant amount of redundant information, as each vertex appears as many times as the number of faces it belongs to.

To resolve this issue, we propose Adjacent Mesh Tokenization (AMT) to obtain a more compact and well-structured $\text{Seq}_V$ than previous method. Our key observation is that the main redundancy of $\text{Seq}_V$ comes from representing each face with three vertices as in~\eqref{Face2V}. This results in vertices that have already been visited appearing redundantly in $\text{Seq}_V$. Therefore, AMT aims to represent each face using only a single vertex whenever possible. As shown in Fig.~\ref{fig:amt} and Algo.~\ref{alg:AMT}, AMT efficiently encodes adjacent faces during tokenization, using only one additional vertex. When no adjacent face is available, as illustrated in the last step of Fig.~\ref{fig:amt}, AMT inserts a special token \& into the sequence to denote this event and restarts the process from a face that has not yet been encoded. To detokenize, simply reverse the tokenization algorithm as described in Algo.~\ref{alg:AMT}.

In the ideal case, where the special token "\&" is rarely used, AMT can reduce the length of $\text{Seq}_V$ obtained by previous methods to nearly one-third. Of course, in extreme cases, such as when each face in the mesh is completely disconnected from others, AMT performs worse than previous methods. However, since the datasets~\cite{deitke2023objaverse, chang2015shapenet} used for AM generation are created by human artists, the meshes generally have well-structured topologies. Thus, the overall performance of AMT is significantly better than previous methods. As shown in Sec.~\ref{quan}, on the Objaverse test set, AMT can reduce the length of $\text{Seq}_V$ by half on average.

\noindent\textbf{Vertices Swap.} Considering two faces $f_1$ and $f_2$: 
$$f_1 = (v_1, v_2, v_3), f_2 = (v_1, v_3, v_4).$$ 

These two faces are connected by edge $(v_1, v_3)$. Assuming we first represent $f_1$ as $(v_1, v_2, v_3)$, AMT will be interrupted because $f_2$ does not contain $v_2$, even though $f_1$ and $f_2$ are actually adjacent. To solve this issue, we introduce a special token \$ to swap vertices. When the \$ token appears in front of a vertex, it indicates that the next face will be formed by the first and the last vertex of the previous face, instead of the last two. For example, the token sequence $(v_1, v_2, v_3, \$, v_4)$ implies the mesh consists of two faces: $(v_1, v_2, v_3)$ and $(v_1, v_3, v_4)$, i.e. $f_1$ and $f_2$.

When swap is enabled, AMT can explore more adjacent faces when searching for the next face. Specifically, in line 13 of Algo.~\ref{alg:AMT}, $AdjacentVertices$ can be modified to additionally include vertices adjacent to the last and third-to-last vertices in $Seq$.

Although the swap operation introduces an additional special token, it reduces the number of interruptions and effectively shortens the token sequence. While adding a special token may potentially increase the difficulty of sequence learning, our experiments in Sec.~\ref{quan} show that it has no noticeable impact.

\noindent\textbf{Discussion on Sorting in Mesh Tokenization.} Both previous methods and AMT initially sort the vertices and faces of the mesh. The primary goal is to process the mesh data into a sequence with a fixed pattern, making it easier for the learning of the sequence model. In AMT, to maintain this design, we consistently choose the face with the earlier index in the sorted list whenever there are multiple choices. Besides, thanks to this design, the token sequence processed by AMT is unique for each mesh. Additionally, AMT prioritizes visiting adjacent faces whenever possible. In contrast, previous methods naively follow the sorted order, often resulting in token sequences where spatially distant vertices are adjacent in the sequence, potentially increasing sequence complexity. As shown in Sec.~\ref{quan}, compared to previous methods, AMT demonstrates significant advantages in both speed and memory usage, as well as improved accuracy, proving that the sequences generated by AMT are more compact and well-structured. 

\noindent\textbf{Discussion on AMT and the use of VQ-VAE.} After obtaining $\text{Seq}_V$, mesh generation methods then need to process it into a token sequence for sequence learning. \cite{siddiqui2023meshgpt} propose to train a VQ-VAE~\cite{van2017neural} to achieve this. They take $\text{Seq}_V$ as input and learn a vocabulary of geometric embeddings with the VQ-VAE. After training the VQ-VAE, they then use the VQ-VAE's quantized features as the input for the transformer~\cite{vaswani2017attention}. Very recently,~\cite{chen2024meshxl} proposed another method to process $\text{Seq}_V$ into a token sequence. They discard VQ-VAE and directly use the discretized coordinates of the vertex as the token index. It is important to emphasize that whether or not VQ-VAE is used does not affect AMT's effectiveness. This is because AMT operates before the aforementioned methods. For example, in the case of using VQ-VAE, AMT first shortens the $\text{Seq}_V$ that represents $\mathcal{M}$, and the shortened $\text{Seq}_V$ is then quantized into an embedding sequence with the VQ-VAE.

\subsection{MeshAnything V2}
\label{sec:V2}
In this section, we introduce MeshAnything V2. It is equipped with AMT and scales up its maximum generated face count from $800$ to $1600$. Without increasing the number of parameters, MeshAnything V2 achieves shape conditioned Artist-Created Mesh (AM) generation with significantly better performance and efficiency. We also use it as an example to demonstrate how AMT can be applied to mesh generation.

Following \cite{chen2024meshanything}, MeshAnything V2 also targets generating AMs aligned to a given shape, allowing integration with various 3D asset production pipelines to achieve highly controllable AM generation. That is, we aim to learn the distribution: \( p(\mathcal{M} | \mathcal{S}) \), where \(\mathcal{M}\) represents the AM and \(\mathcal{S}\) represents the 3D shape condition.

As in \cite{chen2024meshanything}, V2 uses point clouds as the shape condition input \(\mathcal{S}\). We also use the same point cloud–Artist-Created Mesh data pairs \((\mathcal{M}, \mathcal{S})\) collected in~\cite{chen2024meshanything}. The target distribution \( p(\mathcal{M} | \mathcal{S}) \) is learned with a decoder-only transformer with the same size and architecture as in~\cite{chen2024meshanything}. To inject \(\mathcal{S}\) into the transformer, we first encode it with a pretrained point cloud encoder~\cite{zhao2024michelangelo} into a fixed-length token sequence \(\mathcal{T}_S\) and then set it as the prefix of the transformer's token sequence. We then process paired \(\mathcal{M}\) into mesh token sequence \(\mathcal{T}_M\). It is concatenated to the point cloud token sequence as the transformer's ground truth sequence. After training the transformer with cross-entropy loss, we input \(\mathcal{T}_S\) and let the transformer autoregressively generate the corresponding \(\mathcal{T}_M\), which is then detokenized into \(\mathcal{M}\).

The key difference between~\cite{chen2024meshanything} and our method is the way we obtain \(\mathcal{T}_M\). Instead of the naive mesh tokenization method used in~\cite{chen2024meshanything}, we process \(\mathcal{M}\) with the newly proposed Adjacent Mesh Tokenization (AMT) and obtain a more compact and efficient sequence \(\text{Seq}_V\). Following \cite{chen2024meshxl}, we discard the VQ-VAE and directly use the discretized coordinates from \(\text{Seq}_V\) as token indices. We then add a newly initialized codebook entry to represent the \& in the AMT sequence. Finally, we sequentially combine the coordinate token sequence and the special token for \& to obtain the mesh token sequence \(\mathcal{T}_M\) for transformer input. As mentioned in Sec.~\ref{sec:AMT}, it is worth noting that whether or not VQ-VAE is used does not affect the application and effectiveness of AMT.

\begin{table*}[h]
\caption{\textbf{Ablation Study on AMT.} We compare MeshAnything V2 with its variant without AMT. Please refer to Section~\ref{quan} for detailed explanation.}

\centering
\begin{tabular}{lcccccccc}
\hline
\textbf{Method} & \textbf{CD$\downarrow$} & \textbf{ECD$\downarrow$} & \textbf{NC$\uparrow$} & \textbf{\#V$\downarrow$} & \textbf{\#F$\downarrow$} & \textbf{V\_Ratio$\downarrow$} & \textbf{F\_Ratio$\downarrow$} &
\textbf{S\_Ratio$\downarrow$} \\
 & \small($\times 10^{-2}$) & \small($\times 10^{-2}$) & 
 &  &  &  &  \\

\hline
V2 W/O AMT & 0.895 & 4.832 & 0.924 & \textbf{302.4} & \textbf{556.7} & \textbf{1.105} & \textbf{1.062} & 1.000
\\
V2  & \textbf{0.874} & \textbf{4.721} & \textbf{0.933} & 308.6 & 571.8 & 1.127 & 1.097 & \textbf{0.497}\\

\hline
\end{tabular}
\vspace{-3mm}
\label{table:quan}
\end{table*}

To facilitate the transformer's learning of the sequence patterns of AMT, in addition to the absolute positional encoding used in \cite{chen2024meshanything}, we add the following embeddings for Adjacent Mesh Tokenization (AMT): when representing a face with three vertices, we add a specific embedding for the three new vertices; when representing a face with one vertex, we add a different embedding for the single new vertex. Additionally, we provide a distinct embedding for the \& token.

\noindent\textbf{Face Count Condition.} Considering that some applications require control over the approximate face count, we explored the addition of a face count condition in mesh generation. We initialized an embedding book with a size equal to the maximum allowed number of faces. Based on the current face count in the mesh, we retrieve the corresponding embedding from this book and place it after the point cloud prefix to represent the face count. During training, we add a random variable to the face count to introduce some variation and prevent overfitting to an exact condition. During training, we add a random variable to the face count to introduce some variation and prevent overfitting to an exact condition. Additionally, we drop this condition with a 10\% probability to further improve condition robustness.

\noindent\textbf{Masking Invalid Predictions.} PolyGen~\cite{nash2020polygen} introduces Masking Invalid Predictions during inference. In PolyGen, vertices are generated first, followed by faces, and both vertices and faces are sorted. Without applying constraints during inference, generative models may produce results that do not follow the sequence structure, such as generating faces before completing all vertices or generating vertices that violate the sorted coordinates order. PolyGen~\cite{nash2020polygen} addresses this issue by masking invalid logits during the inference phase, ensuring that only valid results are generated.

We followed this design in our PolyGen experiments. Due to the widespread use of coordinate sorting in meshes, this design can also be adopted in other tokenization methods. In the AMT experiments, we incorporated this design as well. For example, we enforced a rule in AMT where, when starting a new strip, at least three vertices must be generated before allowing any interruptions.

\section{Experiments}
\subsection{Implementation Details}
\label{exp: implement}

The main experimental setting of MeshAnything V2 remains consistent with~\cite{chen2024meshanything}, except we equipped it with the newly proposed Adjacent Mesh Tokenization (AMT). We still use OPT-$350\text{M}$~\cite{zhang2022opt} as our autoregressive transformer~\cite{vaswani2017attention} and the pretrained point encoder from~\cite{zhao2024michelangelo}. 

For dataset preparation, we adopt a similar dataset preparation technique as used in~\cite{chen2024meshanything}, with several modifications. Firstly, we noticed that Objaverse contains some duplicate mesh data, where the meshes are identical in shape but differ in texture. We detected these duplicates by calculating the differences in vertex coordinates and retained only one instance of each mesh. Additionally, in~\cite{chen2024meshanything}, the ground truth meshes were first processed into SDF using mesh2sdf~\cite{wang2022dual}. While mesh2sdf occasionally produces some failure cases, we filtered out these failures using Chamfer Distance.
Different from the dataset used in~\cite{chen2024meshanything}, we used meshes with fewer than $1600$ faces as the experimental dataset, compared to $800$ in the previous work, and also included a small portion of data from ObjaverseXL. This resulted in a dataset containing $230$K point clouds and mesh pairs. We randomly sampled $4$K data samples as the evaluation dataset.

To accommodate meshes with more faces, we sampled $8192$ points instead of $4096$ for each point cloud. Besides, unlike~\cite{chen2024meshanything}, we update the point encoder from~\cite{zhao2024michelangelo} during training because we find its accuracy insufficient for handling complex meshes with up to $1600$ faces.

MeshAnything V2 is trained with $32$ A800 GPUs for four days. The batch size per GPU is $8$, resulting in a total batch size of $256$.

\subsection{Qualitative Experiments}

We present the qualitative results of MeshAnything V2. As shown in Fig.~\ref{fig:teaser}, MeshAnything V2 effectively generates high quality Artist-Created Mesh aligned to given shapes. When integrated with various 3D assets production pipelines, V2 successfully achieves highly contrabllable AM generation.

\begin{table*}[h]
\caption{\textbf{Comparison of Tokenization Methods.} We compare various mesh tokenization methods and their impact on the generated mesh results in mesh generation. Note that the metrics are calculated by sampling 10K points cloud on each mesh.}
\centering
\begin{tabular}{lccccccccc}
\hline
\textbf{Method} & \textbf{CD$\downarrow$} & \textbf{ECD$\downarrow$} & \textbf{NC$\uparrow$} & \textbf{\#V$\downarrow$} & \textbf{\#F$\downarrow$} & \textbf{V\_Ratio$\downarrow$} & \textbf{F\_Ratio$\downarrow$} & \textbf{S\_Ratio$\downarrow$} & \textbf{Perplexity$\downarrow$} \\
 & \small($\times 10^{-2}$) & \small($\times 10^{-2}$) &  &  &  &  &  &  &  \\

\hline
Baseline & 2.478 & \textbf{18.21} & 0.893 & 95.4 & 178.9 & 0.974 & 0.940 & 1.000 & \textbf{1.150} \\
Unsort & 8.151 & 31.86 & 0.794 & 117.4 & 213.1 & 1.189 & 1.219 & 1.000 & 1.234 \\
PolyGen(AMT) & 3.226 & 22.97 & 0.872 & \textbf{93.2} & \textbf{172.2} & \textbf{0.928} & \textbf{0.936} & \textbf{0.372} & 1.589 \\
AMT & \textbf{2.348} & 19.33 & 0.904 & 102.7 & 187.1 & 1.013 & 0.983 & 0.492 & 1.363 \\
AMT(Swap) & 2.517 & 19.86 & \textbf{0.913} & 110.8 & 198.9 & 1.098 & 1.102 & 0.455 & 1.416 \\

\hline
\end{tabular}
\vspace{-3mm}
\label{table:tokenization_comparison}
\end{table*}

\subsection{Quantitative Experiments}
\label{quan}

\textbf{Evaluation Metrics.} We follow the evaluation metric settings of~\cite{chen2024meshanything} to quantitatively assess mesh quality. We uniformly sample point clouds from both the ground truth meshes and the generated meshes and compute the following metrics on our $2$K evaluation dataset:
\begin{itemize}
    \item \textbf{Chamfer Distance (CD):} Evaluates the overall quality of a reconstructed mesh by computing chamfer distance between point clouds.
    \item \textbf{Edge Chamfer Distance (ECD):} Assesses the preservation of sharp edges by sampling points near sharp edges and corners.
    \item \textbf{Normal Consistency (NC):} Evaluates the quality of the surface normals.
    \item \textbf{Number of Mesh Vertices (\#V):} Counts the vertices in the mesh.
    \item \textbf{Number of Mesh Faces (\#F):} Counts the faces in the mesh.
    \item \textbf{Vertex Ratio (V\_Ratio):} The ratio of the estimated number of vertices to the ground truth number of vertices.
    \item \textbf{Face Ratio (F\_Ratio):} The ratio of the estimated number of faces to the ground truth number of faces.
    \item \textbf{Perplexity:}~We report training perplexity to measure how well the mesh generation model predicts a sample. A lower perplexity indicates more accurate predictions.
    \item \textbf{Sequence Ratio (S\_Ratio):}~The sequence length compression ratio of the used mesh tokenization methods.

\end{itemize}

\noindent\textbf{Ablation Study.} We ablate the effectiveness of AMT by comparing the results of MeshAnything V2 with its variant without AMT. The variant follows exactly the same settings as V2, except that AMT is replaced with naive mesh tokenization as in previous methods~\cite{siddiqui2023meshgpt,chen2024meshxl,chen2024meshanything}. This also serves as a fairer comparison with \cite{chen2024meshanything}, as the original model of~\cite{chen2024meshanything} was trained on data with fewer than $800$ faces rather than $1600$. Moreover, \cite{chen2024meshanything} is trained on a small total batch size of $64$ while we observed a noticeable performance improvement when the batch size was increased to V2's $256$. We sample 100K points on each mesh to calculate the aforementioned metrics.

As shown in Table~\ref{table:quan}, our results indicate that V2 significantly outperforms its variant, demonstrating the effectiveness of AMT. It shows that AMT not only improves training speed and reduces memory pressure but also enhances generation quality. Notably, although V2 and its variant were trained for the same number of iterations, the variant consumed nearly two times the GPU hours of V2. Additionally, the table shows that AMT takes slightly more vertices and faces. A likely reason for this is that the variant's performance is still too weak, and it tends to ignore details and use a simpler topology when representing complex meshes with high face counts. We also observe that both AMT and the variant have vertex and face ratios greater than $1.0$, meaning they use more faces on average relative to the ground truth, unlike the results in \cite{chen2024meshanything}, which were less than $1$ (around $0.88$). We suspect this is because the dataset for V2 contains more complex meshes with over $800$ faces, which causes the model to occasionally produce more complex topology for simple shapes.

\noindent\textbf{Comparison of Tokenization Methods.} Below, we list the methods to be compared in the following sections, along with their specific implementation details. In the tokenization methods comparison experiments, we use the smaller OPT-125M~\cite{zhang2022opt} and select meshes with up to 400 faces from the aforementioned datasets as the experimental dataset. We sample 10K points on each mesh to calculate the aforementioned metrics.

\begin{itemize}
    \item \textbf{Baseline.} This refers to the naive mesh tokenization method used in previous works, where each face is strictly represented by three vertices.
    
    \item \textbf{PolyGen(AMT).} This is a combination of the tokenization method from PolyGen~\cite{nash2020polygen} with AMT. Polygen first generates vertex coordinates and then expresses the faces by generating vertex indices. We discuss this setting in detail in the Supplementary Materials. 
    \item \textbf{AMT.} This refers to the Adjacent Mesh Tokenization proposed in Sec.~\ref{sec:AMT}, as shown in Figure Aug1.
    \item \textbf{AMT(Swap).} As described in Sec.~\ref{sec:AMT}, this method combines the AMT algorithm with vertices swap. An additional special token is used to represent the swap operation.
    \item \textbf{Unsort.} This is based on the naive mesh tokenization method from previous works~\cite{nash2020polygen,siddiqui2023meshgpt}, but without sorting the mesh. Instead, the mesh is used in the order in which it is stored. It is important to note that the meshes in Objaverse~\cite{deitke2023objaverse} are created by human, and their storage may already include some level of inherent order. The purpose of this method is to compare with the \textbf{Baseline} and verify the importance of sorting.
\end{itemize}

The results are shown in Tab.~\ref{table:tokenization_comparison}. Unsort vs Baseline comparison validates the importance of tokenization method regularity for mesh generation. 

PolyGen(AMT)~\cite{nash2020polygen} achieves the best S\_Ratio (sequence length compression ratio), indicating its high efficiency. However, there is a noticeable performance gap compared to Baseline and AMT, indicating the produced mesh quality is not high enough. The primary reason for this might be that the autoregressive model struggles to interact accurately with previously generated vertices when producing faces, which increases the difficulty of sequence learning.

The results from AMT demonstrates that it shortens the token length without compromising mesh quality, as confirmed by its comparison with the Baseline. Additionally, in AMT(Swap), the introduction of the Swap special token further increases the compression ratio without affecting mesh quality.

From the above experiments, we can conclude that both the regularity of the token sequence and the compression ratio are crucial. Although PolyGen~\cite{nash2020polygen} achieves a better compression ratio, its token sequence organization is not well-suited for sequence learning, making it a sub-optimal mesh tokenization method compared to AMT. In contrast, both AMT and AMT(Swap) reduce the compression ratio without increasing the difficulty of sequence learning, making them more suitable for mesh generation.

\section{Conclusion}

In this work, we present MeshAnything V2, a shape-conditioned Artist-Created Mesh (AM) generation model that generates AM aligned to given shapes. V2 significantly outperfroms MeshAnything~\cite{chen2024meshanything} in both performance and
efficiency with our newly proposed Adjacent Mesh Tokenization (AMT). Different from previous methods that use three vertices to represent a face, AMT uses a single vertex whenever possible. Our experiments demonstrate that AMT averagely reduces
the token sequence length by half. The compact, and well-structured token sequence from AMT greatly enhances sequence learning, thereby significantly improving the efficiency and
performance of AM generation.

{
    \small
    \bibliographystyle{ieeenat_fullname}
    \bibliography{main}
}

\clearpage
\appendix
\section{Additional Experiments}
We provide additional quantitative experiments and detailed discussion about Polygen~\cite{nash2020polygen} tokenization method here.

\subsection{Experiments on Face Count Condition.} We tested the effectiveness of the face count condition by scaling the ground truth face count value by a scalar. We used the V2 model trained on the mesh dataset with fewer than 1600 faces as the test model and sampled 10K point clouds for each mesh to calculate the metrics. During inference, for each ground truth mesh, we input the scaled face count and the paired point cloud into the model and measured the effect on face count and mesh quality. As shown in Tab.~\ref{tab:face}, at a scale ratio of 0.8, the model significantly reduced the face count while maintaining mesh quality. However, at a scale ratio of 0.6, the face count did not decrease further, indicating that while the model has some ability to follow the face condition, it prioritizes mesh quality when the condition becomes difficult to meet. Similarly, when the scale ratio is set to greater than 1, the model exhibits similar behavior.

\subsection{User Study.} We conducted a user study on mesh generation quality to compare MeshAnything~\cite{chen2024meshanything} and our works. For a fair comparison, we randomly sampled 30 meshes with fewer than 800 faces from the evaluation dataset and input their paired point clouds into each model. Users were asked to select the result they preferred from the two options. We collected responses from a total of 43 users, and the voting rates for~\cite{chen2024meshanything} and MeshAnything V2 were 32.2\% and 67.8\%, respectively. This indicates that the results generated by V2 are more aligned with human preference.

\subsection{Discussion on PolyGen Tokenization Method.} Polygen\cite{nash2020polygen} introduces a mesh tokenization approach that differs from other existing methods~\cite{siddiqui2023meshgpt} for mesh generation. It first employs an autoregressive vertex model to generate the 3D coordinates of the mesh's vertices. These vertices are then used as a prefix and fed into another autoregressive face model, which connects these vertices into faces, thus constructing the entire mesh. Since the 3D coordinates are already provided by the vertex model, the face model does not need to estimate the 3D coordinates, but only specifies the connections using vertex indices, significantly reducing the sequence length. During the face generation, this tokenization method can be combined with AMT further shorten the token sequence length.

\begin{table}[h]
\caption{\textbf{Experiments on Face Count Condition.} We control the face count condition using the scale ratio, where 1.0 indicates using the ground truth face count as the condition. The experiments show that our face count condition has the ability to control the number of faces.}
\centering
\begin{tabular}{cccc}
\hline
\textbf{Scale Ratio} & \textbf{CD$\downarrow$} \small($\times 10^{-2}$) & \textbf{V\_Ratio} & \textbf{F\_Ratio} \\
\hline
1.0 & 1.768 & 1.127  & 1.097 \\
0.8 & \textbf{1.734} & 0.928 & 0.912 \\
0.6 & 1.822 & \textbf{0.902} & \textbf{0.882} \\
1.2 & 1.814 & 1.282 & 1.248 \\
1.4 & 1.920 & 1.271 & 1.252 \\
\hline
\end{tabular}
\label{tab:face}
\end{table}

By using vertex indices to represent faces, PolyGen~\cite{nash2020polygen} tokenization consumes only one token to define a face, whereas other methods require three tokens to represent a single vertex. Assuming a vertex is referenced \(n\) times, PolyGen's tokenization requires \(3+n\) tokens, whereas other tokenization methods would require \(3 \times n\) tokens. Although PolyGen spends one additional token when a vertex is referenced only once, in most cases, each vertex is referenced multiple times, allowing PolyGen's approach to save sequence length.

Although PolyGen tokenization effectively reduces the token sequence length, this generation method requires the model to accurately predict vertex positions first, which may not be ideal for an autoregressive model.

In the experiments section of our main paper, we compare Polygen tokenization method with other methods. To make a fair comparison, we merge the two stage generation process of PolyGen~\cite{nash2020polygen} into a single model that first generates vertex coordinates and then expresses the faces by generating vertex indices. Besides, we combine PolyGen's face generation stage with AMT to further reduce the token sequence length. 

\section{Limitations.}

Although there is a large improvement over V1, the accuracy of MeshAnything V2 is still insufficient for industrial applications. More efforts are needed to improve the model's stability and accuracy.

\end{document}